# Unsupervised Deep Feature Extraction for Remote Sensing Image Classification

Adriana Romero, Carlo Gatta and Gustau Camps-Valls, *Senior Member, IEEE*

**Abstract**

This paper introduces the use of *single layer* and *deep* convolutional networks for remote sensing data analysis. Direct application to multi- and hyper-spectral imagery of supervised (shallow or deep) convolutional networks is very challenging given the high input data dimensionality and the relatively small amount of available labeled data. Therefore, we propose the use of *greedy layer-wise unsupervised pre-training* coupled with a highly efficient algorithm for unsupervised learning of sparse features. The algorithm is rooted on *sparse representations* and enforces both population and lifetime sparsity of the extracted features, simultaneously. We successfully illustrate the expressive power of the extracted representations in several scenarios: classification of aerial scenes, as well as land-use classification in very high resolution (VHR), or land-cover classification from multi- and hyper-spectral images. The proposed algorithm clearly outperforms standard Principal Component Analysis (PCA) and its kernel counterpart (kPCA), as well as current state-of-the-art algorithms of aerial classification, while being extremely computationally efficient at learning representations of data. Results show that single layer convolutional networks can extract powerful discriminative features *only* when the receptive field accounts for neighboring pixels, and are preferred when the classification requires high resolution and detailed results. However, deep architectures significantly outperform single layers variants, capturing increasing levels of abstraction and complexity throughout the feature hierarchy.

## I. INTRODUCTION

Earth observation (EO) through remote sensing techniques is a research field where a huge variety of physical signals is measured from instruments on-board space and airborne platforms. A wide diversity of sensor characteristics is nowadays available, ranging from medium and very high resolution (VHR) multispectral imagery to hyperspectral images that sample the electromagnetic spectrum with high detail. These myriad of sensors serve to particularly different objectives, focusing either on obtaining quantitative measurements and estimations of geo-bio-physical variables, or on the identification of materials by the analysis of the acquired images [1]–[3]. Among all the different products that can be obtained from the acquired images, classification maps[1] are perhaps the most relevant ones. The remote sensing image classification problem is very challenging and ubiquitous because land cover and land use maps are mandatory in multi-temporal studies and constitute useful inputs to other processes.

Despite the high number of advanced, robust and accurate existing classifiers [4], the field faces very important challenges:

[1]In the remote sensing community, the term 'classification' is often preferred to the term 'semantic segmentation'. We use the term 'classification' to classify full images in the first application of aerial image classification, and to describe the process of attributing each pixel (or segment) to a single semantic class in subsequent applications.

1) *Complex statistical characteristics of images.* The statistical properties of the acquired images place important difficulties for automatic classifiers. The analysis of these images turns out to be very challenging, especially because of the high dimensionality of the pixels, the specific noise and uncertainty sources observed, the high spatial and spectral redundancy and collinearity, and their potentially non-linear nature[2]. Beyond these well-known data characteristics, we should highlight that spatial and spectral redundancy also suggest that the acquired signal may be better described in *sparse* representation spaces, as recently reported in [4], [6]–[8].

2) *High computational problems involved.* We are witnessing the advent of a *Big Data Era*, especially in remote sensing data processing. The upcoming constellations of satellite sensors will acquire a large variety of heterogeneous images of different spatial, spectral, angular and temporal resolutions. In fact, we are witnessing an ever increasing amount of data gathered with current and upcoming EO satellite missions, from multispectral sensors like Landsat-8 [9], to VHR sensors like WorldView-III [10], the super-spectral Copernicus' Sentinel-2 [11] and Sentinel-3 missions [12], as well as the planned EnMAP [13], HyspIRI [14] and ESA's candidate FLEX [15] imaging spectrometer missions. This data flux will require computationally efficient classification techniques. The current state-of-the-art Support Vector Machine (SVM) [16], [17] is not, however, able to cope with more than some few thousands of labeled data points.

A very convenient way to alleviate the above-mentioned problems is to extract relevant, potentially useful, non-redundant, non-linear features from images in order to facilitate the subsequent classification step. The extracted features could be fed into a simple, cost-effective (ideally linear) classifier. The bottleneck would then be the *feature learning* step. Learning expressive *spatial-spectral features* from hyperspectral images in an *efficient* way is thus of paramount relevance. In addition, and very importantly, learning such features in an *unsupervised* fashion has also become extremely relevant given the few labeled pixels typically available.

## A. Background

Given the typically high dimensionality of remote sensing data, feature extraction techniques have been widely used in the literature to reduce the data dimensionality. While the classical Principal Component Analysis (PCA) [18] is still one of the most popular choices, a plethora of non-linear dimensionality reduction methods, manifold learning and dictionary learning algorithms have been introduced in the last decade.

State-of-the-art *manifold learning* methods [19] include: local approaches for the description of remote sensing image manifolds [20]; kernel-based and spectral decompositions that learn mappings optimizing for maximum variance, correlation, entropy, or minimum noise fraction [21]; neural networks that generalize PCA to encode non-linear data structures via autoassociative/autoencoding networks [22]; as well as projection pursuit approaches leading to convenient Gaussian domains [23]. In remote sensing, autoencoders have been widely used [24]–[27]. However, a number of (critical) free parameters are to be tuned; regularization is an important issue, which is mainly addressed by limiting the network's structure heuristically; and only shallow structures are considered mainly due to the limitations on computational resources and efficiency of the training algorithms. On top of this, very often, autoencoders employ only the spectral information, and in the best of the cases, spatial information is naively included through stacking hand-crafted spatial features.

To authors' knowledge, there is few evidence of the good performance of deep architectures in remote sensing image classification: [28] introduces a deep learning algorithm for classification of (low-dimensional) VHR images; [29] explores the robustness of deep networks to noisy class labels for aerial image classification; and [30] introduces hybrid Deep Neural Networks to enable the extraction of variable-scale features to detect vehicles in satellite images; [31] proposes a hybrid framework based on Stacked

---

[2]Factors such as multi-scattering in the acquisition process, heterogeneities at subpixel level, as well as atmospheric and geometric distortions lead to distinct non-linear feature relations, since pixels lie in high dimensional curved manifolds [4], [5].

Auto-Encoders for classification of hyper-spectral data. Although deep learning methods can cope with the difficulties of non-linear spatial-spectral image analysis, the issues of sparsity in the feature representation and efficiency of training algorithms are not obvious in state-of-the-art frameworks.

In recent years, *dictionary learning* has emerged as an efficient way to learn sparse image features in unsupervised settings, which are eventually used for image classification and object recognition: discriminative dictionaries have been proposed for spatial-spectral sparse-representation and image classification [32], sparse kernel networks have been recently introduced for classification [33], sparse representations over learned dictionaries for image pansharpening [34], saliency-based codes for segmentation [35], [36], sparse bag-of-words codes for automatic target detection [37], and unsupervised learning of sparse features for aerial image classification [38]. These methods describe the input images in sparse representation spaces but do not take advantage of the high non-linear nature of deep architectures.

Therefore, in the context of remote sensing, *unsupervised* learning of features in a *deep* convolutional neural network architectures seeking *sparse* representations has not been approached so far.

## B. Contributions

In this paper, we aim to address the two main challenges in the field of remote sensing data. Therefore, we introduce the use of *deep convolutional* networks for remote sensing data analysis [39] trained by means of an unsupervised learning method seeking *sparse* feature representations. On one hand, (1) *deep* architectures have a highly non-linear nature that is well suited to cope with the difficulties of non-linear spatial-spectral image analysis; (2) *convolutional* architectures only capture local interactions, making them well suited when the input share similar statistics at all location, i.e. when there is high redundancy; (3) *sparse* features are supposed to be convenient to describe remote sensing images [4], [6]–[8]. On the other hand, we want to train deep convolutional architectures efficiently to alleviate the high computational problems involved in remote sensing. Given the typically few labeled data, applying *unsupervised* learning algorithms to train deep architectures is a paramount aspect of remote sensing.

We propose the combination of *greedy layer-wise unsupervised pre-training* [40]–[43] coupled with the highly efficient Enforcing Lifetime and Population Sparsity (EPLS) algorithm [44] for *unsupervised* learning of *sparse* features and show the applicability and potential of the method to extract hierarchical (i.e. deep) sparse feature representations of remote sensing images. The EPLS seeks a sparse representation of the input data (remote sensing images) and allows to train systems with large numbers of input channels efficiently (and numerous filters/parameters), *without requiring any meta-parameter tuning*. Thus, deep convolutional networks are trained *efficiently* in an unsupervised greedy layer-wise fashion [40]–[43] using the EPLS algorithm [44] to learn the network filters. The learned hierarchical representations of the input remote sensing images are used for image/pixel classification, where lower layers extract low-level features and higher layers exhibit more abstract and complex representations.

To our knowledge, this is the first work dealing with *sparse unsupervised deep convolutional networks* in remote sensing data analysis in a systematic way. We want to emphasize the fact that the methodology presented here is fully *unsupervised*, which is a different (and more challenging) setting to the common supervised use of convolutional nets. The main contributions of this paper are

1) *Deep Convolutional Architectures trained with EPLS*. We exploit the properties of the EPLS and extend the work in [44] from single to deep architectures, and from classification of images to semantic segmentation of high-dimensional images, which certainly is a more interesting problem in the field of remote sensing image processing.
2) *Application of the proposed method to very high resolution (VHR), multispectral (MS) and hyperspectral (HS) images*. Unlike [44], which only focused on tiny RGB images, we deal with very high resolution (VHR), multispectral (MS) and hyperspectral (HS) images as well. Moreover, we analyze the influence of deep architectures' meta-parameters on the method's performance.

The rest of the paper is organized as follows. Section II introduces the main characteristics of the proposed algorithm for unsupervised hierarchical (deep) sparse feature extraction: we describe the (deep) convolutional neural network architecture, detail the layer-wise pre-training algorithm, and summarize the unsupervised EPLS algorithm. Section III compares the proposed algorithm to state-of-the-art algorithms in terms of classification accuracy and their expressive power in four different applications: classification of aerial scenes, as well as land-cover classification in VHR, multi- and hyper-spectral images. After a detailed analysis of the results, we end the paper with some concluding remarks and outline of the future work in Section IV.

## II. Unsupervised deep feature learning of remote sensing images

This section introduces the concepts and strategies employed to learn deep features for remote sensing. In II-A we briefly explain the main blocks of a deep convolutional neural network; in II-B we outline a strategy to learn the filters of each layer called *greedy layer-wise unsupervised pre-training* [40], [42], [43]; finally, in section II-C we introduce the EPLS algorithm [44], which is the unsupervised learning strategy employed to learn the network parameters.

### A. Deep Convolutional Neural Networks

Deep neural networks are models that capture hierarchical representations of data. These models are based on the sequential application of a computation "module", where the output of the previous module is the input to the next one; these modules are called *layers*. Each layer provides one representation level. Layers are parameterized by a set of weights connecting input units to output units and a set of biases. In the case of *Convolutional* Neural Networks (CNN), weights are shared locally, i.e. the same weights are applied at every location of the input. The weights connected to the same output unit form a *filter*.

CNN layers consist of: (1) a convolution of the input with a set of learnable filters to extract local features; (2) a point-wise non-linearity, e.g. the logistic function, to allow deep architectures to learn non-linear representations of the input data; and (3) a pooling operator, which aggregates the statistics of the features at nearby locations, to reduce the computational cost (by reducing the spatial size of the image), while providing a local translational invariance in the previously extracted features. Fig. 1 shows an example of CNN, with $L$ layers stacked together. The last convolutional layer is followed by a fully-connected output layer.

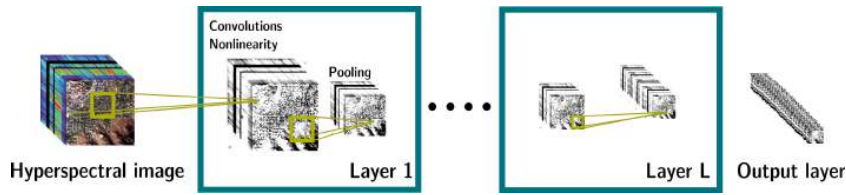

Fig. 1. A graphical representation of a deep convolutional architecture.

The operations performed in a single convolutional layer can be summarized as

$$\mathbf{O}^l = \text{pool}_P(\sigma(\mathbf{O}^{l-1} \star \mathbf{W}^l + \mathbf{b}^l)) \quad (1)$$

where $\mathbf{O}^{l-1}$ is the input feature map to the $l$-th layer; $\theta^l = \{\mathbf{W}^l, \mathbf{b}^l\}$ is the set of learnable parameters (weights and biases) of the layer, $\sigma(\cdot)$ is the point-wise non-linearity, *pool* is a subsampling operation, $P$ is the size of the pooling region[3], and the symbol $\star$ denotes linear convolution. Note that in the context of CNN, the convolution is multi-dimensional with each filter. The input of the first layer is the input data, in this case a multi/hyper-spectral image, i.e. $\mathbf{O}^0 = \mathbf{I}$, where $\mathbf{I} \in \Re^{R^0 \times C^0 \times N_h^0}$ is the input image, $R^0$ and $C^0$ are its width and height and $N_h^0$ is the number of spectral channels (bands). More generally, the input

---
[3]The pooling region is usually square, in this case formed by $P \times P$ pixels.

                                                                                                                                                                  5

to a subsequent layer $l$ is a feature map $\mathbf{O}^{l-1} \in \Re^{R^{l-1} \times C^{l-1} \times N_h^{l-1}}$, where $R^{l-1}$ and $C^{l-1}$ are the width and height of the $l$-th layer's input feature map and $N_h^{l-1}$ is the number of outputs of the $(l-1)$-th layer.

CNN architectures have a significant number of meta-parameters. The most relevant ones may be: (1) the number of layers; (2) the number of outputs per layer; (3) the size of the filters, also called receptive field; and (4) the size and type of spatial pooling.

Another important aspect is *how* to train such architectures. Deep convolutional networks can be trained in a supervised fashion, e.g. by means of standard back-propagation [45]–[47], or in an unsupervised fashion, by means of greedy layer-wise pre-training [40], [42], [43]. Unsupervised greedy layer-wise pre-training has been successfully used in the literature [40], [42], [43], [48], [49] to train deep CNN. Supervised methods usually require a large amount of reliable labeled data, which is difficult to obtain in remote sensing classification problems. Therefore, in the case of multi- and hyper-spectral images, it is preferred to use an unsupervised learning strategy given the typically few available labeled pixels per class.

### B. The greedy layer-wise unsupervised pre-training strategy

Greedy layer-wise unsupervised pre-training [40], [42], [43] is based on the idea that a local (layer-wise) unsupervised criterion can be applied to pre-train the network's parameters, allowing the use of large amounts of unlabeled data. After pre-training, the network's parameters are set to a potentially good local minima, from which supervised learning (called *fine-tuning*) can follow. However, deep networks have also been trained in a purely unsupervised way, skipping the fine-tuning step [48]. Patch-based training is the most commonly used approach to learn the convolutional layers' parameters by means of unsupervised criteria [50]. It consists in using a set of randomly extracted patches from input images (or feature maps) to train each layer. After that, the layer weights are applied to each input location to obtain output convolutional feature maps that will serve as input to the next layer.

Algorithm 1 shows the pseudo-code of a greedy layer-wise unsupervised pre-training strategy, as introduced in [40], [42], [43]. The algorithm expects as input a set images $\mathcal{D}^0 = \{\mathbf{O}_i^0\}_{\forall i}$ and a deep architecture with $L$ layers. Then, it trains each layer of the deep architecture in a patch-based fashion and provides as output the parameters of all layers $\{\theta^1, \theta^2, \ldots, \theta^L\}$, i.e. the (pre)-trained deep architecture with $\theta^l = \{\mathbf{W}^l, \mathbf{b}^l\}$ and $l \in \{1, 2, \cdots, L\}$. For each layer $l$ (line 1), the algorithm extracts $N$ random patches from the feature maps (or images) in $D^{l-1}$ to generate $\mathbf{H}^{l-1} \in \mathcal{R}^{N \times N_h^{l-1}}$ (line 2). Each row of $\mathbf{H}^{l-1}$ corresponds to a vectorized patch and each column represents an input dimension. After that, it learns the layer's parameters $\theta^l$ applying an $UnsupervisedCriterion$ on $\mathbf{H}^{l-1}$ (line 3). In our case, the unsupervised criterion is the EPLS algorithm, which is detailed in subsection II-C. Then, the set of output feature maps $\mathcal{D}^l = \{\mathbf{O}_i^l\}_{\forall i}$ of the trained layer $l$ is computed from the set of input feature maps $\mathcal{D}^{l-1} = \{\mathbf{O}_i^{l-1}\}_{\forall i}$ by performing feature extraction (see Section II-D for more details) (line 4). The new set of feature maps $\mathcal{D}^l$ is subsequently used to train the next layer. The same procedure is repeated for each layer until $l = L$.

---
**Algorithm 1** Greedy layer-wise unsupervised pre-training

    **Input:** $\mathcal{D}^0, L$
    **Output:** $\{\theta^1, \theta^2, \ldots, \theta^L\}$, where $\theta^l = \{\mathbf{W}^l, \mathbf{b}^l\} \forall l \in \{1, 2, ..., L\}$
1: **for** $l = 1 \to L$ **do**
2:    Generate $\mathbf{H}^{l-1} \in \mathcal{R}^{N \times N_h^{l-1}}$ by randomly extracting $N$ patches from $\mathbf{O}_i^{l-1} \in \mathcal{D}^{l-1}$
3:    $\theta^l \leftarrow UnsupervisedCriterion(\mathbf{H}^{l-1})$
4:    $\mathcal{D}^l = \{\mathbf{O}_i^l : FeatureExtraction(\mathbf{O}_i^{l-1}, \theta^l), \quad \forall \mathbf{O}_i^{l-1} \in \mathcal{D}^{l-1}\}$ (see Eq. 1)
5: **end for**

---

### C. Unsupervised learning criteria with sparsity

Sparsity is among the properties of a good feature representation [48], [50]–[53]. Sparsity can be defined in terms of *population sparsity* and *lifetime sparsity*. On one hand, population sparsity ensures simple representations of the data by allowing only a small subsets of outputs to be active at the same time [54]. On the other hand, lifetime sparsity controls the frequency of activation of each output throughout the dataset, ensuring rare but high activation of each output [54]. State-of-the-art unsupervised learning methods such as sparse Restricted Boltzmann Machines (RBM) [40], Sparse Auto-Encoders (SAE) [53], Sparse Coding (SC) [52], Predictive Sparse Decomposition (PSD) [49], Sparse Filtering [48] and Orthogonal Matching Pursuit (OMP-$k$) [55] have been successfully used in the literature to extract sparse feature representations. OMP-$k$ and SC seek population sparsity, whereas SAE seek lifetime sparsity. OMP-$k$ trains a set of filters by iteratively selecting an output of the code to be made non-zero in order to minimize the residual reconstruction error, until at most $k$ outputs have been selected. The method achieves a sparse representation of the input data in terms of *population sparsity*. SAE train the set of filters by minimizing the reconstruction error while ensuring similar activation statistics through all training samples among all outputs, thus ensuring a sparse representation of the data in terms of *lifetime sparsity*. However, the great majority of these methods have numerous meta-parameters and/or enforce sparsity at the expense of adding meta-parameters to tune.

In [44], we introduced EPLS, a novel, meta-parameter free, off-the-shelf and simple algorithm for unsupervised sparse feature learning. The method provides discriminative features that can be very useful for classification as they capture relevant spatial and spectral image features jointly. The method iteratively builds a sparse target from the output of a layer and optimizes for that specific target to learn the filters. The sparse target is defined such that it ensures both population and lifetime sparsity. Figure 2 summarizes the steps of the method in [44]. Essentially, given a matrix of input patches to train layer $l$, $\mathbf{H}^{l-1}$, we need to: (1) compute the output of the patches $\mathbf{H}^l$ by applying the learned weights and biases to the input, and subsequently the non-linearity; (2) call the EPLS algorithm to generate a sparse target $\mathbf{T}^l$ from the output of the layer, such that it ensures population and lifetime sparsity; and (3) optimize the parameters of the layer (weights and biases) by minimizing the $L_2$ norm of the difference between the layer's output and the EPLS sparse target:

$$\theta^{l*} = \arg\min_{\theta^l} ||\mathbf{H}^l - \mathbf{T}^l||_2^2 \qquad (2)$$

The optimization is performed by means of an out-of-the-box mini-batch Stochastic Gradient Descent (SGD) with adaptive learning rates [56]. From now on, we will use the superscript $b$ to refer to the data related to a mini-batch, e.g. the output of a layer $\mathbf{H}^l \in \mathcal{R}^{N \times N_h^l}$ will now be $\mathbf{H}^{l,b} \in \mathcal{R}^{N_b \times N_h^l}$, where $N_b < N$ is the number of patches in a mini-batch.

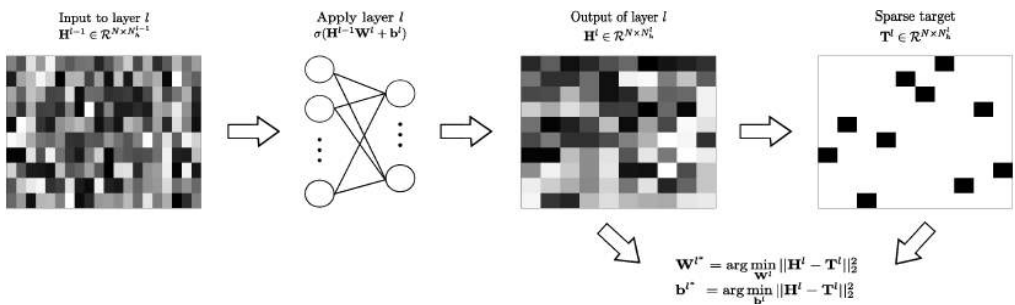

Fig. 2. Illustration of how EPLS generates the output target matrix.

Algorithm 2 recapitulates how the EPLS builds the sparse target matrix from the output matrix of a layer. Let $\mathbf{H}^{l,b}$ be the mini-batch output matrix of a layer $l$, composed of $N_b$ output vectors of dimensionality $N_h^l$. Let $\mathbf{T}^{l,b}$ be the sparse target matrix built by the EPLS, with the same dimensions as $\mathbf{H}^{l,b}$. Starting with no activation in $\mathbf{T}^{l,b}$ (line 1) and the output of the system $\mathbf{H}^{l,b}$ normalized between [0,1] (line 2), the algorithm processes a row $\mathbf{h}$ of $\mathbf{H}^{l,b}$ at each iteration (line 4). In line 5, the algorithm selects the output $k$ of the $n$-th row that has the maximal activation value $\mathbf{h}_j$ minus an inhibitor $\mathbf{a}_j$ to be set as one "hot code", thus ensuring population sparsity. The inhibitor $\mathbf{a}_j$ is initialized to zero. It "counts" the number of times an output $j$ has been selected, increasing its inhibition progressively by $\frac{N_h^l}{N}$ until reaching maximal inhibition, where $N$ is the total number of training patches. This prevents the selection of an output that has already been activated $\frac{N}{N_h^l}$ times and, thus ensures lifetime sparsity. In line 6, the $k$-th element of the $n$-th row of the target matrix $\mathbf{T}^{l,b}$ is activated, ensuring population sparsity. In line 7, the inhibitor is updated and, finally, in line 9, the complete output target $\mathbf{T}^{l,b}$ is remapped to the active/inactive values of the corresponding non-linearity. More details on the EPLS algorithm can be found in [44].

**Algorithm 2** EPLS [44]

    **Input:** $\mathbf{H}^{l,b}$, $\mathbf{a}$, N
    **Output:** $\mathbf{T}^{l,b}$, $\mathbf{a}$
1: $\mathbf{T}^{l,b} = 0$
2: $\mathbf{H}^{l,b} = \frac{\mathbf{H}^{l,b} - \min(\mathbf{H}^{l,b})}{\max(\mathbf{H}^{l,b}) - \min(\mathbf{H}^{l,b})}$
3: **for** $n = 1 \to N_b$ **do**
4:     $\mathbf{h}_j = \mathbf{H}^{l,b}_{n,j}$ $\forall j \in \{1, 2, \ldots, N_h^l\}$
5:     $k = \arg\max_j (\mathbf{h}_j - \mathbf{a}_j)$
6:     $\mathbf{T}^{l,b}_{n,k} = 1$
7:     $\mathbf{a}_k = \mathbf{a}_k + \frac{N_h^l}{N}$
8: **end for**
9: Remap $\mathbf{T}^{l,b}$ to active/inactive values

### D. Feature Extraction

After training the parameters of a network, we can proceed to extract feature representations. To do so, we must choose an encoder to map the input feature map of each layer to its representation, i.e we must choose the non-linearity to be used after applying the learned filters to all input locations. A straightforward choice is the use of a natural encoding, i.e. whichever encoding is associated to the training procedure. When using EPLS to train networks, the natural encoding is the non-linearity used to compute the output of each layer. However, different training and encoding strategies can be combined together. Encodings that lead to sparse representations have proven to be effective in the literature, e.g. soft-threshold encoding is a popular choice, which involves a tunable meta-parameter to control the desired degree of sparsity [55].

Moreover, the use of polarity split has shown to further improve the performance of many experiments [55]. Polarity splitting takes into account the positive and negative components of a code in the following way:

$$\begin{aligned}\mathbf{O}^l_+ &= pool_P(\sigma(\mathbf{O}^{l-1} \star \mathbf{W}^l + \mathbf{b}^l)) \\ \mathbf{O}^l_- &= pool_P(\sigma(\mathbf{O}^{l-1} \star (-\mathbf{W}^l) + \mathbf{b}^l)) \\ \mathbf{O}^l &= [\mathbf{O}^l_+, \mathbf{O}^l_-],\end{aligned} \quad (3)$$

where $\mathbf{O}^l$ is the concatenation of the positive and negative components of the code. Polarity split results in doubling the number of outputs and is usually applied to the output layer of the network.

Summarizing, we train deep architectures by means of greedy layer-wise unsupervised pre-training in conjunction with EPLS and choose a feature encoding strategy for each specific problem. Initial parameters are randomly drawn from $\mathcal{N}(0, 10^{-8})$. Each layer is trained for a minimum of 20 epochs and a maximum of $N_h^l$ epochs. If the relative training error decrease between epochs is very small, the training stops. The mini-batch size is initialized to $\frac{N}{N_h^l}$ and, as is standard practice, the mini-batch size is doubled every time the training error between two consecutive epochs increases.

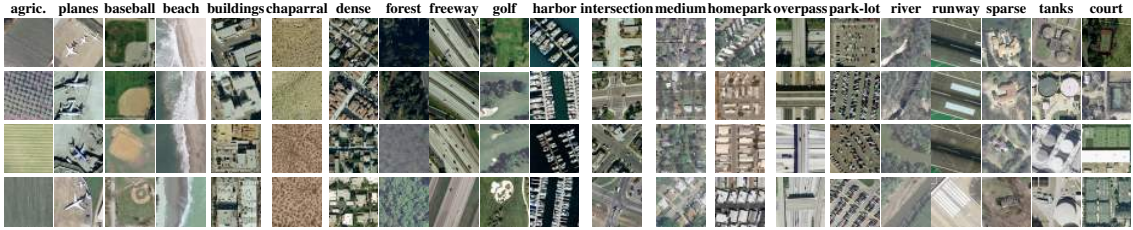

Fig. 3. The ground truth dataset contains 100 images for each 21 land-use classes, from which we show four samples per class.

## III. Experimental results

This section is devoted to illustrate the capabilities of the presented algorithm in different scenarios of image classification and segmentation found in remote sensing. We study relevant issues as the potential of extracted features for land cover/use classification in aerial, VHR, multi- and hyper-spectral images. We study problems with a wide diversity of input data dimensionality, number of classes and amount of available labeled data. Finally, we pay attention to particularly relevant issues when training the proposed method such as the importance of depth and sparsity, the impact of the pooling stages, and the learned hierarchical representations.

### A. Aerial Scene Classification

*1) Data Collection:* We validate the aerial scene classification on the UCMerced dataset. The dataset contains manually extracted images from the USGS National Map Urban Area Imagery collection[4]. UCMerced consists of 256×256 color images from 21 aerial scene categories, with resolution one foot per pixel. The dataset contains highly overlapping classes and has 100 images per class. Figure 3 depicts some images per land-use class of interest.

*2) Experimental setup:* We follow the experimental setup described in [38] and randomly select 80 images per class for training and leave the remaining 20 ones for testing. As in [38], we report the mean accuracy obtained over five runs.

To validate our method, we follow the experimental pipeline of [38]: (1) Extract random patches from raw images and normalize them for contrast and brightness; (2) Train a network in a patch-based fashion by means of an unsupervised criterion, in our case, EPLS with logistic activation; (3) Use the trained network parameters and an encoding strategy to retrieve sparse representations; (4) Pool the upper-most feature map into four quadrants via sum-pooling; and (5) Feed the pooled features to a linear SVM classifier. We tuned the SVM $C$ parameter using 5-fold cross-validation. Following this pipeline, we define four experimental settings to highlight the competitiveness of our method in terms of performance.

In the first setting, we use the same parameters as [38] and set a receptive field of $16 \times 16$ pixels with stride 8 pixels to train a single layer network with $N_h^1 = 1000$, for fair comparison. We train our system on normalized raw image patches by means of EPLS with logistic non-linearity and retrieve the sparse features by applying the network parameters with natural encoding and polarity split. Features are then pooled into four quadrants and are fed into a linear SVM classifier. As reported in [44], we achieved a classification performance of $74.34 \pm 3.0\%$, which is significantly higher than the $62.7 \pm 1.72\%$ reported in [38] when pairing OMP-1 training with soft-threshold encoding (tuned to achieve maximum performance) on normalized raw pixels. Note that when considering the result reported in [38] using OMP-1 with its natural encoding on normalized raw pixels ($13.86 \pm 1.31\%$), the results are even more impressive.

In the second setting, we train single layer networks with varying $N_h^1$. In this case, we use a receptive field of 15×15 pixels with stride 1 pixel. As in the first experiment, we train the systems on normalized raw input image patches by means of EPLS and retrieve the sparse features by applying the network parameters and a natural encoding with polarity split. Again, features are sum-pooled into four quadrants and fed to an SVM. Figure 4 (left) shows the classification performance of our single layer approach (solid red line), compared to the best results reported in [38] (solid green line), for different $N_h^1$ values. As shown in the figure, our method outperforms the method in [38] for all $N_h^1$ in terms of average performance in the 5 runs, while having no training nor encoding meta-parameters to tune. Moreover, [38] requires to train the single layer network on SIFT features to achieve such performance, whereas we train all our networks on raw image patches normalized for contrast and brightness, i.e. we do not require any prior feature extraction.

---

[4]http://vision.ucmerced.edu/datasets/landuse.html

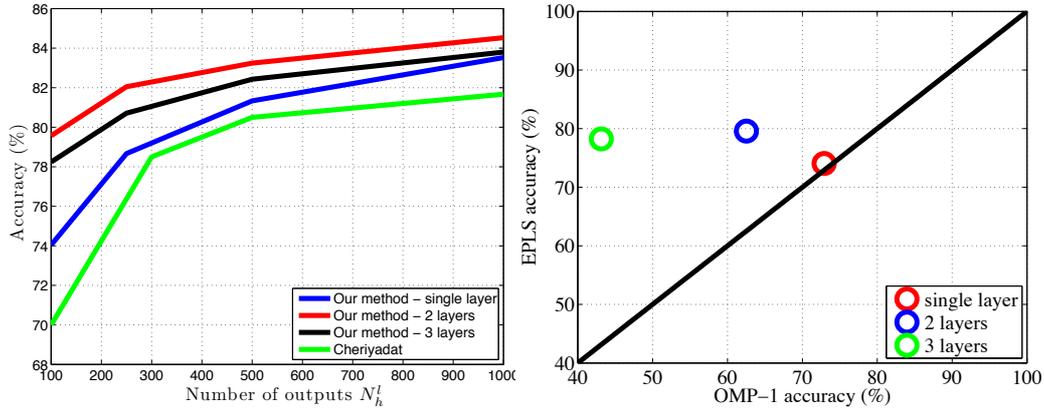

Fig. 4. Left: UCMerced accuracy given different number of outputs and different network architectures (best seen in color). Right: Comparison of EPLS training against OMP-1 for $N_h^l = 100$ on architectures of increasing depth (best seen in color).

To strengthen the results, we report the per class users and producers for the second setting single layer network with $N_h^1 = 1000$ and compare the producer's accuracy to the best results reported in [38], [57]. Figure 5 (left) shows the users and producers obtained by our approach. The proposed method achieves a very high sensitivity and specificity for most of the classes, in particular for chaparral, harbor, parking lot and runway. Errors are mainly from scenes with similar spatial structures, such as buildings and residential areas, for which the database contains three similar subclasses (medium, sparse and dense residential areas). We also compare in Figure 5 (right) our approach to previously reported results [38], [57]. The producers' accuracy is in general favorable for our method: in 14 out of 21 classes, we obtained better results than in [57], and in 15 out of 21 in [38]. These results encourage the use of the method, and the exploitation of combined approaches in future research.

In the third setting, we experimented with deep architectures to further exploit the possibilities of our method. In this case, we used a receptive field of $7 \times 7$ pixels, with a stride of 1 pixel. We trained deep CNN composed of two and three layers, respectively. We used 100,000 normalized patches to train each layer of the network. We applied a non-overlapping max-pooling of $2 \times 2$ pixels at each representation layer, except for the last layer, which divides the output feature map into 4 quadrants and applies sum-pooling for fair comparison with the single layer architectures. We trained each layer by means of EPLS with a logistic non-linearity. We applied a linear encoding (i.e. identity activation function) to each middle layer and a rectifier encoding with polarity split to the last representation layer. The last layer features, once learned, are fed into a linear SVM. As in the second experimental setting, we trained the same architecture with varying $N_h^l$. Note that all these deep configurations are built by stacking layers with the same number of units. Figure 4 (left) shows the classification performance of both the 2-layer CNN (solid blue line) and 3-layer CNN (solid black line), for different $N_h^l$ configurations. As shown in the figure, 2-layer CNN improve the previous single layer results, for all $N_h^l$. The 2-layer CNN with $N_h^l = 500 \; \forall l \in \{1, 2\}$ outperforms the single layer network with $N_h^1 = 1000$, whereas the 2-layer CNN with $N_h^l = 1000 \; \forall l \in \{1, 2\}$ achieves $84.53\%$ accuracy, outperforming all previous results. However, when increasing the number of layers to

 10

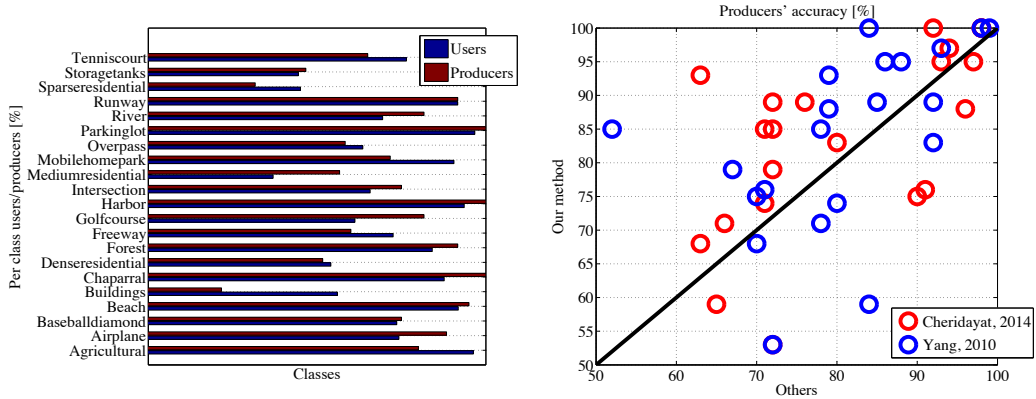

Fig. 5. Results in UCMerced experiment. Left: users and producers per class obtained with the proposed method. Right: comparison of our method to previous works in the literature in terms of producers (accuracy per class) [38], [57]. We took the results for the best algorithms in both [57] (their Figure 8, 'color' model) and [38] (his Fig. 13, dense SIFT descriptors) (best seen in color).

3, the accuracy starts dropping. We argue whether the UCMerced dataset could benefit from a higher level of abstraction in its feature representation, given the (high) amount of texture present in its images. Even if it could benefit from the higher abstraction, the layer-wise pre-training might be too greedy and a fine-tuning step might be required to achieve better performance as we increase the depth of the network. Furthermore, the receptive field size could be appropriately tuned to improve the results; note that the image region considered in the 3rd layer is much bigger than the ones considered by the 1st and 2nd layers. Finally, it is worth noticing that as we increase the number of layers, the number of parameters increases dramatically and the model becomes more prone to overfit.

After highlighting the impact of stacking hierarchical (deep) representations, we designed an experiment to assess the importance of sparsity to achieve good representations. In order to highlight the relevance of lifetime sparsity in deep scenarios, we used OMP-1 as a substitute of EPLS and reproduced the experiments in Figure 4 (left) for $N_h^l = 100$. Figure 4 (right) reports the obtained results. In the case of the single layer network (red circle), EPLS achieves slightly better results than OMP-1. However, as shown in the Figure, OMP-1 seems not to be able to take advantage of depth. When adding a second layer to the architecture (blue circle), OMP-1 experiences a performance drop of $10.38\%$, whereas ELPS improves its performance by $5.52\%$. When adding a third layer to the architecture (green circle), both OMP-1 and EPLS decrease their performances. OMP-1's performance drop is particularly dramatic (from $72.90\%$ in the single layer architecture to $43.14\%$ in the 3-layer architecture). We argue this dramatic performance drop is related to the OMP-1's lack of lifetime sparsity, which makes the algorithm suffer from dead outputs (i.e. outputs that never activate). While increasing the network's depth, the dead outputs' effect becomes more significant and impacts the performance of the method. Therefore, enforcing lifetime sparsity is crucial for EPLS to achieve good performance.

*B. Very high resolution (VHR) image classification*

Very high spatial resolution (VHR) has been one of the major achievements of satellite imagery of the last decades. Sensors providing sub-metric resolution have been developed and satellites such as QuickBird, GeoEye-1 or WorldView-3 have been or are about to be launched. These sensors provide images that are unique in terms of spatial detail and open a wide range of challenges for geospatial information processing. This application example studies how to extract discriminative features from VHR imagery in an unsupervised way via the deep CNN proposed here.

*1) Data Collection:* To do the experiments, we used *two* VHR images acquired with the Quickbird instrument. The satellite data were obtained from Quickbird II, which employs a four-band sensor with

            11

2.4-m spatial resolution for blue, green, red and near-infrared spectral wavelengths and a 0.6-m resolution panchromatic band (©2008 Digitalglobe, all rights reserved). Images acquisitions were obtained to coincide with seasonal base flow conditions and field surveys during late summer or early fall periods. Each image was georeferenced using field-collected ground control points that yielded an average root mean square geolocation error of less than 1.5 m.

The images were acquired over Nayak-Middle Fork (1659×1331×4) of the Flathead River in the Nyack flood plain bordering Glacier National Park, Montana, and the Kol flood plain (1617×1660×4), during 2008. Both images have been widely used to study and characterize the physical complexity of North Pacific Rim rivers to assist wild salmon conservation. The labeled *land used* classes are actually related to such properties: shallow shore, parafluvial, and orthofluvial salmon habitats types [58], [59].

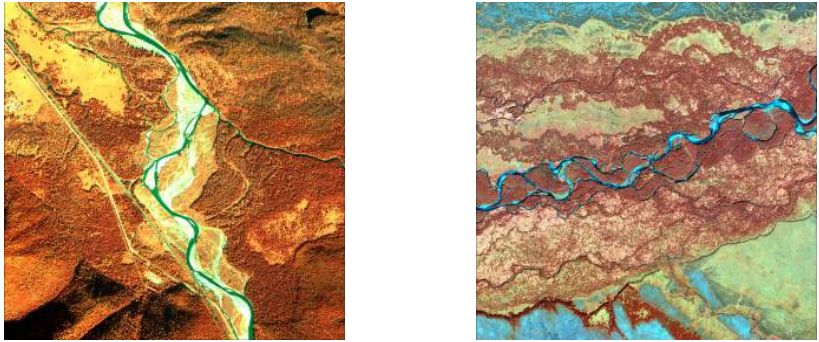

Fig. 6. RGB composition of the two VHR images considered for classification: 'Nayak - Middle Fork' (left) and 'Kol' (right).

*2) Experimental setup:* The setting involves an independent feature extraction and classification per image, which is a standard scenario in remote sensing image classification. The aim of the experiment is to assess in two independent images the capabilities of CNN architectures to extract useful features to capture spectral-spatial structure for habitat classification. The experiments are conducted for a varying number of layers and number of training examples. For the single layer architecture the receptive field is set to $5 \times 5$ pixels, while for deep architectures the receptive field is set to $3 \times 3$ pixels. In all cases, each layer has 200 outputs, i.e. $N_h^l = 200 \ \forall l$. CNN are trained in a layer-wise fashion by means of EPLS with logistic non-linearity as unsupervised criterion. A non-overlapping max-pooling of size $2 \times 2$ pixels is applied to the output of the hidden layers, except for the last representation layer, which does not perform any pooling operation. We choose a natural encoding without polarity split to map the input of each layer to its output representation. The output of the last layer is fed to a 1-Nearest Neighbor (1-NN) classifier with Euclidean distance.

Figure 7 shows the classification results in terms of overall accuracy (OA) and kappa statistic for the two VHR images, as a function of the number of training samples ($\{0.5\%, 1\%, 2.5\%, 5\%, 10\%\}$) and the number of layers (1–6) considered in the model. Three main conclusions are drawn from the experiments: 1) results are improved with more training samples, as expected; 2) non-linear feature extraction (both kPCA and CNN) outperform the linear PCA; and 3) the deeper the network, the better the results. We observed an average gain with a deep CNN of about +10% for the Nayak image and of +20% for the Kol image in terms of kappa statistic. Results saturate for 6 layers and for more than 5% of training samples (this issue has been also observed in the hyper-spectral image classification problem in the following sections).

## C. Multispectral image classification

This experiment is concerned with the challenging problem of cloud screening using multispectral images.

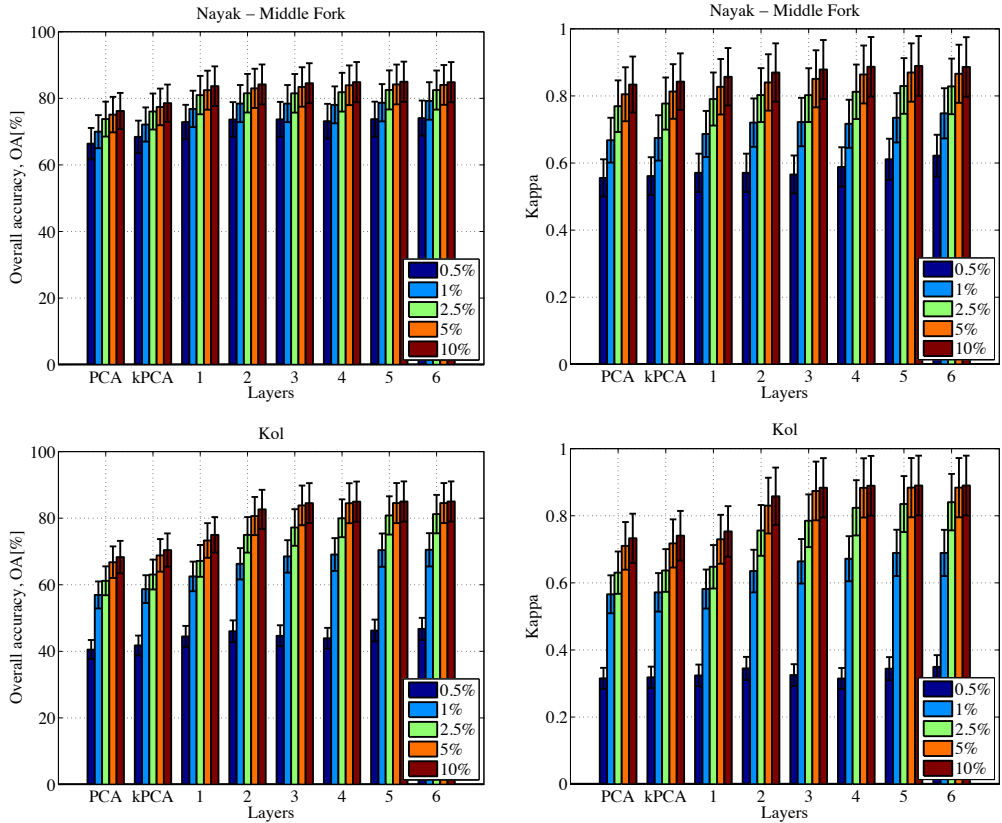

Fig. 7. Classification results (OA and kappa statistic) in the form of the average $\pm$ standard deviation bars over 10 realizations of the classification experiment in two VHR images, as a function of the number of training samples and CNN depth.

*1) Data Collection:* To do the experiments, we used *seven images* acquired with the Medium Resolution Imaging Spectrometer (MERIS) instrument on-board the Environmental Satellite (ENVISAT). In particular, we used images acquired over Abracos (2004), Ascension Island (2005), Azores (2004), Barcelona (2006), Capo Verde (2005), Longyearbyen (2006) and Mongu (2003). All images are of size $321 \times 490$, and all 16 channels were used for the feature extraction. The selected images represent different scenarios extremely useful to validate the performance of the method, including different landscapes; soils covered by vegetation or bare; and critical cases given the special characteristics of the induced problems: ice and snow. Note that the images cover different characteristics as well: geographic location, date and season, type of cloud, and surface types.

*2) Experimental setup:* Most approaches to tackle the problem of cloud screening rely on neural networks or kernel machines trained with supervision: they assume the existence of a labeled training dataset that serves to tune model parameters [60]–[64]. In our approach, more in line of the unsupervised clustering approach presented in [65], we alternatively test the scheme of unsupervised deep feature extraction followed by a simple 1-NN classifier with Euclidean distance. The main goal is to attain a fully unsupervised, meta-parameter free scheme.

The proposed experimental setup allows to assess the expressive power of the extracted features in a very consistent way. Essentially, the scheme performs an *image-fold cross-validation* through both features and samples. Classification of pixels in an image needs to learn the features for that particular image: in our setting, we apply the deep CNN learned from other images without any re-training of the weights or filters. For classification, the inferred features are fed to the 1-NN classifier. The deep architecture was designed such that all layers have the same number of outputs, i.e. $N_h^l = 120 \, \forall l$. We used a receptive field of size $5 \times 5$ pixels. We trained the CNN layer-wise by means of EPLS with logistic activation function[5], and incorporated a max-pooling operator of size $2 \times 2$ after each intermediate representation. We did not perform any kind of pooling after the last representation layer. We applied natural encoding without polarity split to extract the features of each layer. We compare the feature extraction performed by the proposed CNN to the feature extraction performed by PCA and kPCA. For the sake of a fair comparison, we extracted a fixed number of features in all cases: we set a number of features $N_f = 120$ for both PCA and kPCA and used $N_h^l = 120 \; \forall l$ in the CNN case ($l = 2$ exhibited the best results). For kPCA, we used an RBF kernel and set the lengthscale parameter to the average distance between all training samples, as a reasonable estimate (note that the feature extraction is unsupervised, so there are no labels at this stage to tune kernel parameters). We fed the extracted features to a 1-NN classifier.

*3) Expressive and discriminative power:* Given the difficulty of obtaining labeled training samples in this problem, we simulate the expressive power of the extracted features using different numbers of training pixels for classification. Figure 8 shows the evolution of two classification scores with the number of training samples: the overall accuracy (OA) and the estimated Cohen's kappa statistic, $\kappa$. Results show that both measures are consistent (convergence with the number of samples), and no big differences are observed between OA and $\kappa$ so the classifications are, in principle, unbiased. We noted that in general the proposed CNN performs better than the other feature extractors, especially when few training samples are used for classification. This suggests that the extracted features are more discriminative and rich, which can be compensated with the information conveyed by using more labeled examples. Interestingly, we found that tuning the kernel parameter for kPCA becomes very complex; even though we here show the results for the standard 'rule of thumb' of setting the lengthscale to the average distance between points, we tried different other alternatives but still the simple spectral approach (no feature extraction) yielded better results than kPCA.

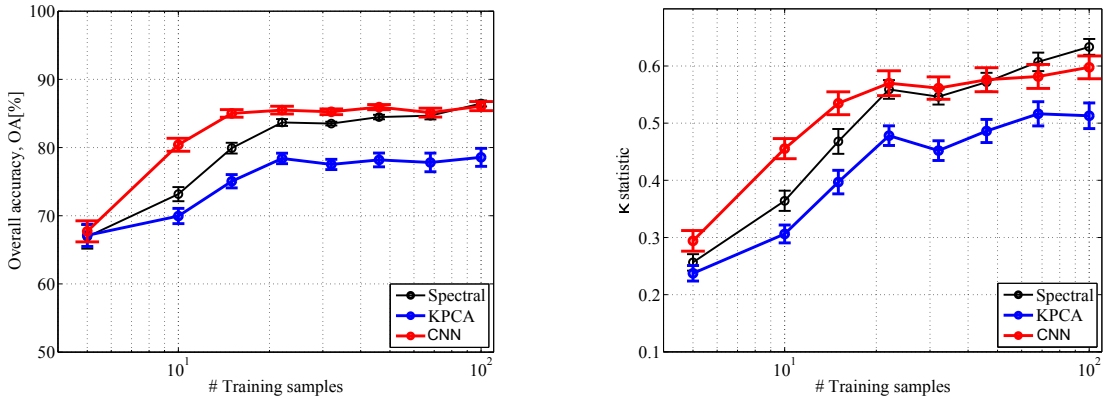

Fig. 8. Classification results (mean and standard deviation) over the seven MS images used as a function of the number of training samples.

*4) Visual Validation of the results:* Figure 9 shows the classification maps obtained with a 1-NN classifier using the spectral information only, and on top of the extracted features using kPCA and CNN. In terms of numerical results, CNN clearly outperforms the rest of approaches. In all cases, we observed an average gain between 5-30% in $\kappa$ statistic (see e.g. outstanding results on 'Abracos', 'Mongu' or 'Longyearbyen' scenes). It should be also noted that in some particular cases (e.g. 'Azores', 'Capo Verde', and 'Longyearbyen') CNN does only significantly improve the results over kPCA, not the spectral approach, which is probably due to the low efficiency in extracting spatially relevant features over areas highly affected by clouds over snowy mountains (that lead to non-interesting features), sunglint (as in the east part of Capo Verde) or very easy images as in the case of compact clouds over sea (Azores scene). In some other cases with similar conditions (e.g. sunglint in Ascension island) the gain obtained by the proposed CNN is noticeable (+6% over spectral features and +12% over kPCA), especially because

---

[5]We also tried with a linear activation function but the extracted features were less performing.

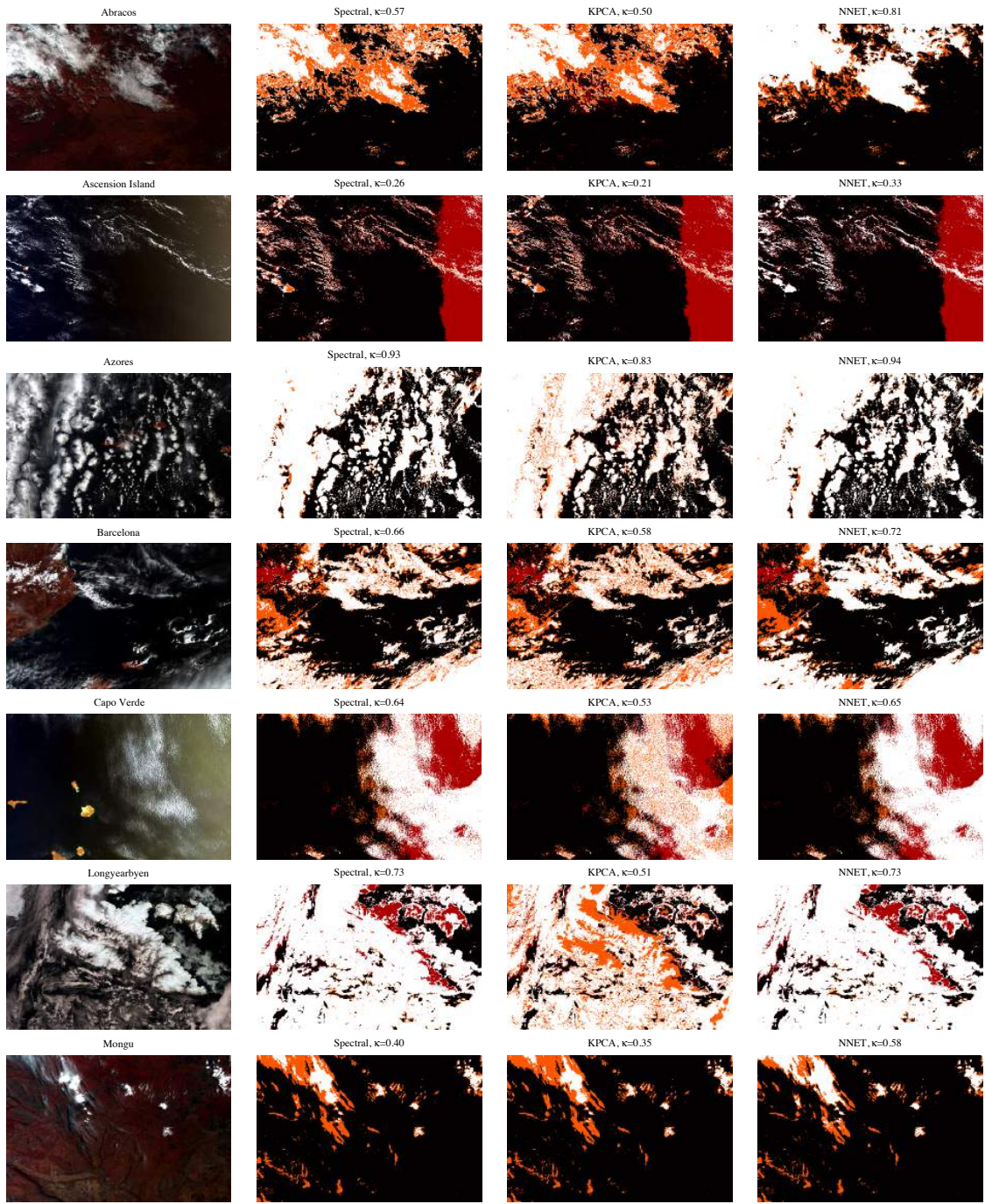

Fig. 9. Classification maps for the different MERIS images obtained by a 1-NN classifier working with pure spectral information only (no feature extraction was done, just use the raw spectral signature for classification), and the extracted features via kPCA and CNN. The obtained $\kappa$ statistic is shown on top of the maps.

                                                                                                                                                                                 15

of the high rate of positive detections in the part of the scene not affected by the sunglint. Another interesting case of study is the Barcelona image in which similar maps are obtained by all methods; nevertheless CNN yields a lower false alarm rate in compact structures (southern and northern big clouds) thus demonstrating that the spatial-spectral information has been very well captured. A more dramatic gain showing this particularity of the method is shown in the Abracos and Mongu scenes, where +30% and 18-23% average gain is attained by CNN, respectively, thanks to a reduced false alarm rate in clouds over flat landscapes.

### D. Hyperspectral image classification

This section illustrates the performance of the proposed method in a challenging hyper-spectral image classification problem. We compare the features extracted by CNN of varying depth to the ones extracted by PCA and kPCA in terms of expressive power, classification accuracy, and robustness to the number of labeled examples.

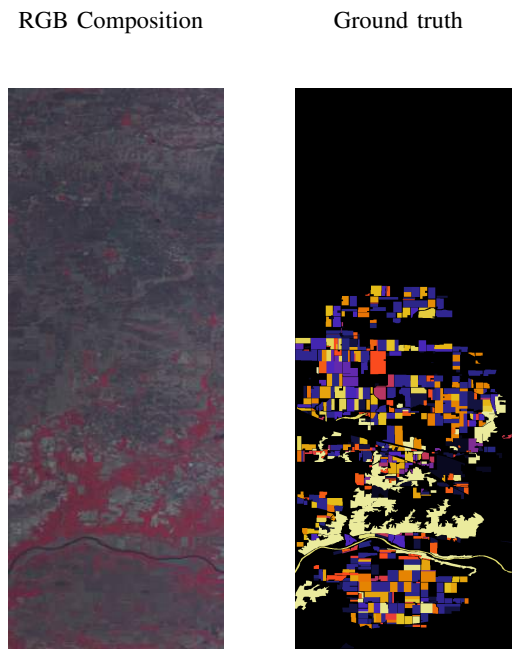

Fig. 10. Color composition (left) and the available reference data (right) for the AVIRIS Indian Pines data set.

*1) Data Collection:* This battery of experiments are conducted on the well-known AVIRIS Indiana's Indian Pines test site acquired in June 1992. A small portion ($145 \times 145$ pixels) of the original image has been extensively used as a benchmark image for comparing classifiers[6]. Here, however, we consider the whole image, which consists of $614 \times 2166$ pixels and 220 spectral bands, with a spatial resolution of 20 m. This data set represents a very challenging land-cover classification scenario.

From the $58$ different land-cover classes available in the original ground truth, we discarded $20$ classes since an insufficient number of training samples were available[7], and thus, this fact would dismiss the planned experimental analysis. The background pixels were not considered for classification purposes. We also removed 20 bands that are noisy or covering the region of water absorption, finally working with 200 spectral bands. See Fig. 10 for a RGB composite and the labeled ground truth of the image.

[6]ftp://ftp.ecn.purdue.edu/biehl/MultiSpec/92AV3C.lan
[7]i.e., less than 1000 samples

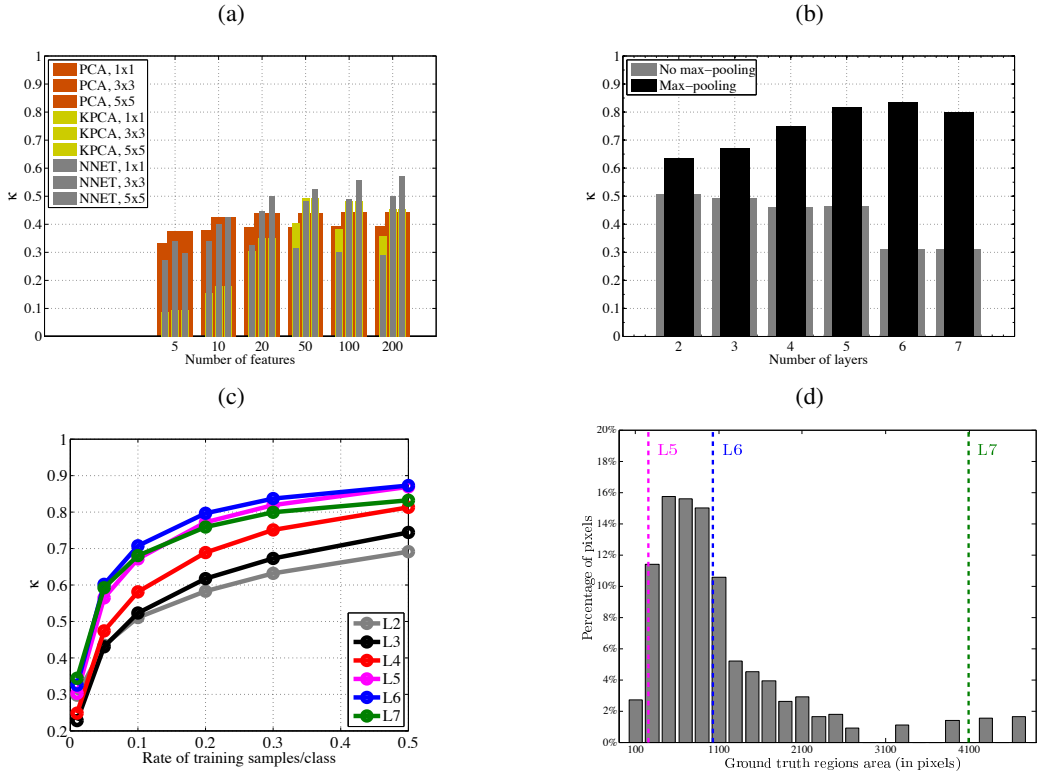

Fig. 11. Classification accuracy estimated with the kappa statistic for (a) several numbers of features, spatial extent of the receptive fields (for the single layer network) or the included Gaussian filtered features (for PCA and kPCA) using 30% of data for training; (b) impact of the number of layers on the networks with and without pooling stages; (c) for different rates of training samples, $\{1\%, 5\%, 10\%, 20\%, 30\%, 50\%\}$, with pooling; and (d) percentage of ground truth pixels as a function of labeled region areas (see text for details).

*2) Experimental setup:* We extract different numbers of features $N_f = \{5, 10, 20, 50, 100, 200\}$ by means of PCA and kPCA and design CNN architectures of varying depth with the same number of outputs per layer $N_h^l = \{5, 10, 20, 50, 100, 200\}$ $\forall l$. We evaluate the feature extraction in all scenarios varying the rates of training samples per class, $\{1\%, 5\%, 10\%, 20\%, 30\%, 50\%\}$. For each deep architecture, we chose different receptive field sizes ($1 \times 1$, $3 \times 3$ or $5 \times 5$), using the same receptive field for all layers, to study the relevance of spatial information. Moreover, we train the layers both with and without a max-pooling operation to assess the effect of the downscaling factor. We train CNN in a layer-wise fashion by means of EPLS with logistic non-linearity. We then use a natural encoding without polarity split to extract the network's features. For kPCA, we use an RBF kernel and set the lengthscale parameter to the average distance between all training samples. In all cases, we feed extracted features to a 1-NN classifier with Euclidean distance, which is used to measure the performance of each system by means of the estimated Cohen's kappa statistic, $\kappa$, in an independent test set made of all remaining examples.

*3) Expressive and discriminative power:* Figure 11(a) shows the $\kappa$ statistic for several numbers of extracted features ($N_f$ and $N_h^1$, respectively) using PCA, kPCA and single layer networks. Both kPCA and the networks yield poor results when a low number of features are extracted, and drastically improve their performance for more than 50 features. Single layer networks stick around $\kappa = 0.3$ for pixel-wise classification, even with increased number of features. Nevertheless, there is a relevant gain when spatial information is considered. The best results are obtained for $N_h^1 = 200$ features and 5×5 receptive fields. With these encouraging results, we decided to train deeper CNN using 30% of the available training samples per class and $N_h^l = 200$ output features per layer. Results with and without the max-pooling operation are shown in Fig. 11(b). Two main conclusions can be drawn: first, deeper networks improve the accuracy enormously (the 6-layer network reaches the highest accuracy of $\kappa = 0.84$), and second, including the max-pooling operation after each intermediate layer revealed to be extremely beneficial. We should stress that this result clearly outperforms the previously reported state-of-the-art result ($\kappa = 0.75$) obtained with a SVM on the same experimental setting [66].

*4) Robustness w.r.t. number of training labels:* Another question to be addressed is the robustness of the features in terms of number of training examples. Figure 11(c) highlights that using a few supervised samples to train a deep CNN can provide better results than using far more supervised samples to train a single layer one. Note, for instance, that the 6-layer network using 5% samples/class outperforms the best single layer network using 30% of the samples/class.

*5) Need and limitation of spatial pooling:* Special attention should be devoted to the 7-layer network. In this case, the accuracy decreases since the potential contribution of an additional layer is strongly counterbalanced by the heavily reduced spatial resolution of the additional max-pooling. The topmost layer has no max-pooling since it is used as output. To corroborate this explanation, we created the histogram in Figure 11(d), which shows the percentage of ground truth pixels as a function of labeled region areas. As it can also be seen in Figure 10(right), the labeled regions are mainly rectangular with an average area around 500 pixels. Vertical lines in Figure 11(d) show the theoretical spatial resolution in the case the output layer is resized using a nearest neighbor interpolation. As it can be observed, when using 7 layers (L7, green), the resolution is too low to capture regions smaller than 4096 pixels (64×64). It has to be noted that we perform the upscaling of the output layer by means of a bilinear interpolation; this explains why, despite the lower spatial resolution, the result using 6 layers is still superior to the one with 5 layers.

*6) Learned features:* An important aspect of the proposed deep architectures lies in the fact that they typically give rise to compact hierarchical representations. The best three features extracted by the networks according to the mutual information with labels are depicted in Fig. 12 for a subset of the whole image. It is worth stressing that the deeper we go, the more complicated and abstract features we retrieve, except for the seventh layer that provides spatially over-regularized features due to the downscaling impact of the max-pooling stages. Interestingly, it is also observed that, the deeper structures we use, the higher spatial decorrelation of the best features we obtain.

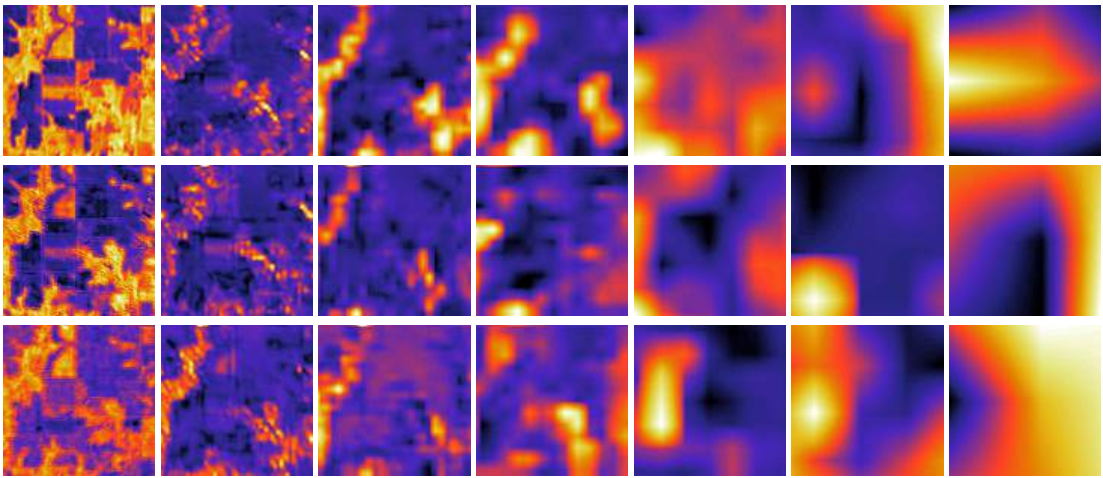

Fig. 12. Best three features (in rows) according to the mutual information with the labels for the outputs of the different layers 1st to 7th (in columns) for a subregion of the whole image.

## IV. CONCLUSIONS

We introduced deep learning for unsupervised feature extraction of remote sensing images. The proposed approach consists of using a convolutional neural network trained with an unsupervised algorithm that promotes two types of feature sparsity: population and lifetime sparsity. The algorithm trains the network parameters to learn hierarchical sparse representations of the input images that can be fed to a simple classifier. We should stress that the unsupervised learning of features is computationally very efficient, having a computational cost equal to OMP-1 while clearly outperforming it. Furthermore, the feature extraction stage is meta-parameter free, whereas the classification stage involves either one or zero free-parameters. Note that we applied the linear SVM classifier with one tunable parameter in experiment Section III-A for fair comparison with the state-of-the-art. We trained *deep* convolutional networks in a greedy layer-wise fashion and performed experiments to analyze the influence of depth and pooling of such networks on a wide variety of remote sensing images of different spatial and spectral resolutions, from multi- and hyper-spectral images, to very high geometrical resolution problems.

Results reveal that the trained networks are very effective at encoding spatio-spectral information of the images. Experiments showed that (1) including spatial information is essential in order to avoid poor performance in single layer networks; (2) combining high numbers of output features and max-pooling steps in deep architectures is crucial to achieve excellent results; and (3) adding new layers to the deep architecture improves the classification score substantially, until the repeated max-pooling steps heavily reduce the features spatial resolution and/or the number of parameters becomes too large, thus inducing a form of overfitting.

Further work is tied to assessing generalization of the encoded features in multi-temporal and multi-angular image settings, as well as to explore the suitability of the extracted features to perform biophysical parameter retrieval. Moreover, it would be interesting to analyze the feature's degree of sparsity required at each layer to achieve a discriminative system w.r.t. the classifier and adapt the EPLS to train towards the desired degree of sparsity at each layer. Furthermore, to avoid overfitting while increasing the number of parameters, algorithms such as drop-out [67] could be tested. Finally, since greedy layer-wise pre-training has shown to require a supervised finetuning step to make the network parameters more task specific, and thus improve the network's performance, further investigation should be devoted to find alternatives to train deep networks without relying on large amounts of labeled data, or even the more challenging task of being able to cope with completely unsupervised settings. Given the large amounts of processing required, algorithms should aim at remaining computationally efficient, especially at training time.

## Acknowledgments

The authors wish to thank Antonio Plaza from the University of Extremadura, Spain, for kindly providing the AVIRIS dataset, and Prof. Diane Whited at the University of Montana for the VHR imagery used in some experiments of this paper.

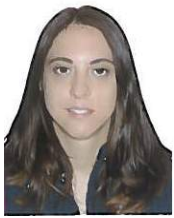

**Adriana Romero** Adriana Romero received the degree in Computer Engineering in 2010 from Universitat Autònoma de Barcelona, and the Master degree in Artificial Intelligence from Universitat Politècnica de Catalunya in 2012. She is currently a PhD candidate at Universitat de Barcelona, working on assisting the training of very deep neural networks. Her main research interests revolve around unsupervised and supervised deep learning and computer vision. More broadly, she is interested in the process of learning in areas that can deepen the understanding of the human nature, such as perception.

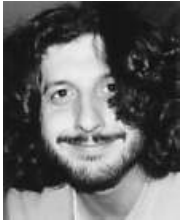

**Carlo Gatta** Carlo Gatta obtained the degree in Electronic Engineering in 2001 from the Università degli Studi di Brescia (Italy). In 2006 he received the Ph.D. in Computer Science at the Università degli Studi di Milano (Italy), with a thesis on perceptually based color image processing. In September 2007 he joined the Computer Vision Center at Universitat Automona de Barcelona (UAB) as a postdoc researcher working mainly on medical imaging. He is member of the Computer Vision Center and the BCN Perceptual Computing Lab. He is currently a senior researcher at the Computer Vision Center, under the Ramon y Cajal program. His main research interests are image processing, medical imaging, computer vision, machine learning, contextual learning and unsupervised deep learning.

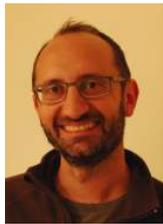

**Gustau Camps-Valls** (M'04, SM'07) received a B.Sc. degree in Physics (1996), in Electronics Engineering (1998), and a Ph.D. degree in Physics (2002) all from the Universitat de València. He is currently an associate professor (hab. Full professor) in the Department of Electronics Engineering. His is a research coordinator in the Image and Signal Processing (ISP) group, http://isp.uv.es. He has been Visiting Researcher at the Remote Sensing Laboratory (Univ. Trento, Italy) in 2002, the Max Planck Institute for Biological Cybernetics (Tübingen, Germany) in 2009, and as Invited Professor at the Laboratory of Geographic Information Systems of the École Polytechnique Fédérale de Lausanne (Lausanne, Switzerland) in 2013. He is interested in the development of machine learning algorithms for geoscience and remote sensing data analysis. He is an author of 120 journal papers, more than 150 conference papers, 20 international book chapters, and editor of the books "Kernel methods in bioengineering, signal and image processing" (IGI, 2007), "Kernel methods for remote sensing data analysis" (Wiley & Sons, 2009), and "Remote Sensing Image Processing" (MC, 2011). He's a co-editor of the forthcoming book "Digital Signal Processing with Kernel Methods" (Wiley & sons, 2015). He holds a Hirsch's $h$ index $h = 40$, entered the ISI list of Highly Cited Researchers in 2011, and Thomson Reuters ScienceWatch® identified one of his papers on kernel-based analysis of hyperspectral images as a Fast Moving Front research. In 2015, he got an ERC consolidator grant on statistical learning for Earth observation data analysis. He is a referee and Program Committee member of many international journals and conferences. Since 2007 he is member of the Data Fusion technical committee of the IEEE GRSS, and since 2009 of the Machine Learning for Signal Processing Technical Committee of the IEEE SPS. He is member of the MTG-IRS Science Team (MIST) of EUMETSAT. He is Associate Editor of the IEEE TRANSACTIONS ON SIGNAL PROCESSING, IEEE SIGNAL PROCESSING LETTERS, and IEEE GEOSCIENCE AND REMOTE SENSING LETTERS. Visit http://www.uv.es/gcamps for more information.